\documentclass{article}
\usepackage[T1]{fontenc}
\usepackage{iclr2027_conference,times}
\usepackage{precipitation_preprint}
\usepackage{amsmath,amssymb,amsfonts,bm,graphicx,booktabs,multirow,microtype}
\usepackage{caption,enumitem}
\usepackage{hyperref,url}
\hypersetup{colorlinks=true,citecolor=blue,linkcolor=black,urlcolor=blue,
pdftitle={Diffusion Precipitation Generation with Medium-range Weather Forecasting},
pdfauthor={Guowen Li; Yang Liu; Mengxuan Chen; Jiashun Cheng; Haoyuan Liang; Shilei Cao; Yujie Wang; Juepeng Zheng; Fang Wang; Yerong Feng; Meng Jin; Hong Cheng; Yutong Lu; Haohuan Fu},
pdfsubject={Observation-aligned daily precipitation generation},
pdfkeywords={precipitation forecasting, diffusion models, weather forecasting}}

\title{\raggedright Diffusion Precipitation Generation with Medium-range Weather Forecasting}
\author{%
\begin{minipage}{\dimexpr\textwidth-2\tabcolsep\relax}\centering\normalfont\small
Guowen Li$^{1,*}$, Yang Liu$^{2,*}$, Mengxuan Chen$^{1}$, Jiashun Cheng$^{3}$, Haoyuan Liang$^{1}$,\\
Shilei Cao$^{4}$, Yujie Wang$^{4}$, Juepeng Zheng$^{4,5,\dagger}$, Fang Wang$^{6}$, Yerong Feng$^{7}$,\\
Meng Jin$^{8}$, Hong Cheng$^{2}$, Yutong Lu$^{4,5}$, Haohuan Fu$^{1,5}$\\[6pt]
{\fontsize{9}{11}\selectfont
$^{1}$Tsinghua University \quad $^{2}$The Chinese University of Hong Kong\\
$^{3}$The Hong Kong University of S \& T \quad $^{4}$Sun Yat-Sen University\\
$^{5}$NSCC-SZ \quad $^{6}$China Meteorological Administration\\
$^{7}$GBA Weather Research Center \quad $^{8}$Huawei Technologies Co., Ltd\\[3pt]
\href{mailto:zhengjp8@mail.sysu.edu.cn}{\texttt{zhengjp8@mail.sysu.edu.cn}}\\[2pt]
$^{*}$Equal contribution. \quad $^{\dagger}$Corresponding author.}
\end{minipage}%
}
\date{}
\begin{document}
\maketitle
\begin{abstract}
Accurate daily precipitation forecasts are important for weather-sensitive decisions. Although recent Weather Foundation Models (WFMs) have achieved significant progress, they are trained on reanalysis data, which exhibit regional and intensity-dependent biases. Since retraining WFMs on observation-based targets is costly, we study \emph{Dynamics--Precipitation Decoupling}, a plug-and-play framework that uses a separately trained generator to map non-precipitation forecasts from a frozen weather model to observation-aligned daily precipitation. Its key challenges are to capture heterogeneous rainfall distributions and align daily accumulations with atmospheric signals that evolve across space and time. To address them, we train a diffusion generator conditional on atmospheric states and satellite-gauge estimates of precipitation. A multi-scale recurrent encoder summarizes atmospheric evolution, while cross-attention aligns atmospheric and precipitation features during denoising. Extensive evaluation of various backbones, including Pangu-Weather, GraphCast, and HRES, on global and regional forecasting shows improved daily precipitation accuracy and event detection, with better representation of localized accumulation patterns. Global ensemble evaluation shows lower Brier scores than ENS and training-from-scratch GenCast at all reported lead times, with lower CRPS in most comparisons. Remarkably, a heavy-rainfall case study using 2,288 stations in China further supports improved daily accumulation patterns and event detection against ground observations.
\end{abstract}

\section{Introduction}

Accurate daily precipitation forecasting is a core task in weather prediction, playing a pivotal role in disaster preparedness, agricultural production, transportation management, and water-resource planning~\citep{hou2014global,vogel2019effects}. Traditional approaches mainly rely on Numerical Weather Prediction (NWP), which solves complex atmospheric dynamical and thermodynamic equations to simulate atmospheric evolution, or statistical methods, which mine relationships between historical precipitation and meteorological factors such as relative humidity~\citep{bauer2015quiet,palmer2017primacy,sun2018review,wilks2011statistical,benziane2024survey}. However, NWP systems are computationally intensive and remain sensitive to uncertainties in initial conditions and physical parameterizations~\citep{bauer2021digital}. These limitations increasingly motivate the development of more data-driven forecasting approaches.
Initially, and in most subsequent precipitation studies, models were trained on observational data, such as radar sequences, and focused on short-term forecasts over local regions, referred to as precipitation nowcasting~\citep{an2025deep,leinonen2023latent,gong2024cascast}. Recent development of Weather Foundation Models (WFMs) trained on ERA5~\citep{hersbach2020era5}, including FourCastNet~\citep{pathak2022fourcastnet} and GenCast~\citep{price2025probabilistic}, has enabled the exploration of global medium-range precipitation forecasting. 

Nevertheless, precipitation in ERA5 is produced through reanalysis model physics and data assimilation, and previous evaluations have documented \textbf{regional and intensity-dependent biases}, including systematic over- or underestimation and distorted spatiotemporal distributions~\citep{xie2022evaluation,jiao2021evaluation}. Consequently, localized precipitation extremes are often smoothed or underrepresented despite accurate large-scale atmospheric forecasts. 
To obtain a more direct observation-aligned precipitation target, we use the \textbf{I}ntegrated \textbf{M}ulti-satellit\textbf{E} \textbf{R}etrievals for \textbf{G}PM (IMERG) satellite-gauge product~\citep{pradhan2022review}. 
A straightforward solution would be to replace the precipitation target and retrain WFMs; however, this is impractical because of the prohibitive training cost. For instance, training GraphCast required about four weeks on 32 Cloud TPU v4 devices~\citep{lam2023learning}. 

Therefore, we study \emph{Dynamics--Precipitation Decoupling}: a separately trained generator maps non-precipitation forecasts from a frozen backbone to observation-aligned daily precipitation. This requires addressing two challenges:
\begin{itemize}[leftmargin=*]
    
    \item Daily precipitation exhibits heterogeneous spatial patterns and skewed accumulation distributions~\citep{liu2025downscaling, gao2022earthformer}, as shown in Figure~\ref{fig:top_figure}(a). Under conditional uncertainty, deterministic models trained with pointwise regression can favor conditional averages, smoothing local accumulation maxima and rainband structures~\citep{ravuri2021skilful}.
    
    \item Atmospheric conditioning poses a nonlocal spatiotemporal alignment challenge. Daily precipitation reflects the cumulative outcome of evolving atmospheric processes, whereas key predictors such as moisture transport and vertical motion may occur earlier and over surrounding or upstream regions~\citep{trenberth2003changing}. As a result, precipitation at a given location cannot be explained by atmospheric conditions at the same grid point and time alone.
\end{itemize}

Recent studies have shown the potential of generative models to preserve localized structures and sharpen precipitation boundaries in radar nowcasting and regional downscaling~\citep{gao2023prediff,yu2024diffcast,gong2024cascast,liu2025downscaling}. In this work, we propose a plug-and-play conditional diffusion generator addresses this challenge through two components, as shown in Figure~\ref{fig:top_figure}(c), which consists of: (1) a multi-scale recurrent encoder captures atmospheric evolution in compact, daily precipitation features; (2) cross-attention aligns atmospheric and precipitation features across scales to guide denoising. The generator learns from paired ERA5 and IMERG precipitation, using only non-precipitation states as conditioning inputs. During inference, its atmospheric conditions are provided by a WFM or NWP system, and a noisy precipitation field is iteratively refined through a reverse-diffusion process under the guidance of encoded atmospheric features. Preliminary experiments in Figure~\ref{fig:top_figure}(b) show that such a model, without backbone-specific retraining, outperforms both the backbone's native precipitation forecasts and forecast-field fine-tuning. Extensive experiments cover three complementary levels: global forecasting at $1.5^\circ$, regional forecasting at $0.25^\circ$, and a heavy-rainfall case study using observations from 2,288 stations in China. Across data-driven and numerical backbones, we evaluate deterministic skill, probabilistic accuracy, input sensitivity, and module efficiency. Station observations further support improved daily accumulation patterns and event detection beyond IMERG-based evaluation. Our main contributions are summarized as follows:

\begin{figure}[t]
    \centering
    
    \includegraphics[width=1\textwidth]{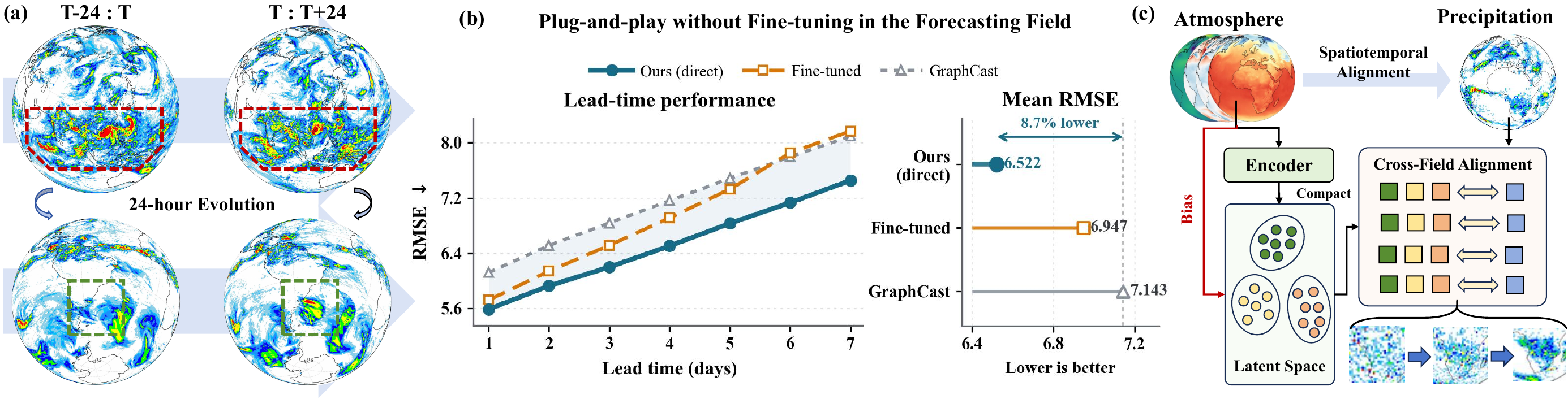}
    
    \caption{Motivation and overview. (a) Consecutive daily precipitation fields exhibit substantial variability. (b) Directly plugging the generator into forecast models outperforms forecast-field fine-tuning. (c) The encoder compresses atmospheric variables into spatiotemporal features, followed by cross-field alignment with precipitation to guide generation. Together, these observations motivate our Dynamics--Precipitation Decoupling paradigm and direct plug-in design for precipitation generation.}
    
    \label{fig:top_figure}
    
\end{figure}

\begin{itemize}[leftmargin=*]
    
    \item We highlight \textbf{Dynamics--Precipitation Decoupling}, separating atmospheric evolution from observation-supervised precipitation generation and enabling IMERG-aligned learning without retraining a complete weather forecasting system.
    \item We develop a forecast-conditioned diffusion architecture that combines a multi-scale atmospheric encoder with an attention-conditioned denoiser to model 24-hour precipitation distributions.
    \item We reuse one ERA5-trained generator with aligned forecasts from frozen Pangu-Weather, GraphCast, and HRES, outperforming available native precipitation forecasts without training a full weather model from scratch.
    \item We provide complementary global, regional, and station-level evaluations of forecast skill and input sensitivity. Validation against 2,288 stations shows improved local rainfall intensity patterns and event detection during the evaluated heavy-rainfall event.
\end{itemize}

\section{Related Work}

\paragraph{Precipitation forecasting with machine learning}
Existing learning-based precipitation forecasting spans two main settings~\citep{gao2021deep,ashok2022systematic,ayzel2020rainnet}. Radar-based nowcasting extrapolates recent observations over regional domains, with representative methods including ConvLSTM~\citep{shi2015convolutional}, Rainformer~\citep{bai2022rainformer}, and NowcastNet~\citep{zhang2023skilful}. These methods capture local precipitation evolution but generally target lead times of only a few hours. WFMs such as FourCastNet~\citep{pathak2022fourcastnet} and GenCast~\citep{price2025probabilistic} extend data-driven forecasting to global domains and medium-range lead times, while FuXi-S2S and ClimaX address subseasonal and climate-oriented tasks~\citep{chen2024machine,nguyen2023climax,watson2022climatebench}. Their precipitation products nevertheless remain tied to system-specific supervision, and documented biases in ERA5 precipitation can limit agreement with satellite--gauge observations~\citep{xie2022evaluation,jiao2021evaluation}. The most related study is NeuralGCM~\citep{yuval2026neural}, which also utilizes IMERG supervision at $2.8^\circ$. However, our work addresses a different setting: rather than integrating precipitation prediction into the forecasting backbone, we decouple precipitation generation from atmospheric forecasting. A separately trained generator maps non-precipitation forecast states to observation-aligned 24-hour precipitation, allowing the same generator to be reused across aligned WFM and NWP backbones without backbone-specific retraining. This design targets medium-range precipitation forecasting at finer spatial resolutions of \(1.5^\circ\) and \(0.25^\circ\), with particular emphasis on localized rainfall structures.

\paragraph{Diffusion models for weather forecasting}
Diffusion-based weather studies cover full-state forecasting, regional atmospheric emulation, and precipitation nowcasting~\citep{bauer2015quiet,tyralis2024review}. GenCast~\citep{price2025probabilistic} generates probabilistic global atmospheric trajectories, while Continuous Ensemble~\citep{andrae2025continuous} produces temporally consistent ensembles at flexible lead times. StormCast~\citep{pathak2026kilometer} and HRRRCast~\citep{abdi2025hrrrcast} emulate convection-allowing regional evolution, whereas PreDiff and DiffCast~\citep{gao2023prediff,yu2024diffcast} generate future radar sequences from recent precipitation observations. These methods therefore evolve complete atmospheric states, emulate a particular regional system, or extrapolate precipitation history. Our framework instead retains an existing WFM or NWP backbone and performs cross-variable generation, using sequences of non-precipitation forecast states to condition an independently IMERG-supervised diffusion generator.

\paragraph{Precipitation post-processing and downscaling}
Precipitation post-processing commonly calibrates or corrects existing precipitation forecasts using observations~\citep{hewson2021low,liu2023deep,tian2024post,zhang2025statistical}. Same-variable downscaling reconstructs finer precipitation fields from coarse satellite or model estimates~\citep{zhu2023towards,zhu2025exploring,ibrahim2026two}. CorrDiff~\citep{mardani2025residual} extends this setting to multivariate downscaling and additional field synthesis through regression and residual diffusion.
These approaches generally transform or refine an existing precipitation field while we generate observation-aligned precipitation directly from non-precipitation atmospheric states, without requiring a precipitation forecast as input. The task is therefore neither precipitation correction nor spatial super-resolution, but observation-supervised cross-variable generation on the conditioning grid.

\section{Background}

\paragraph{Problem definition.}
We study the weather forecasting~\citep{lam2023learning, kochkov2024neural, price2025probabilistic} of $V$ weather parameters at the latitude-longitude grid $\bm{G} \in \mathbb{R}^{H \times W \times 2}$, where $H$ and $W$ are the height and width of the grid that depend on the resolution of latitude and longitude, and $\bm{G}_{h, w, :} = (\lambda_{h}, \phi_{w}) \in \Omega = [-90^{\circ}, 90^{\circ}]\times [-180^{\circ}, 180^{\circ}]$. 
At hour $t$, the global non-precipitation atmospheric state is represented by $\bm{X}_t \in \mathbb{R}^{H \times W \times V}$. Here, $V=V_A C+V_S$, where $V_A$, $V_S$, and $C$ denote the numbers of upper-air variables, surface variables, and pressure levels, respectively.
Given the initial weather conditions $(\bm{G}, \bm{X}_{0})$, our goal is to learn a neural network to forecast 24-hour accumulated precipitation at future time horizons, as shown in the following:
\begin{equation}
    \begin{aligned}
        (\bm{\hat{Y}}_{24}, \bm{\hat{Y}}_{48}, \cdots, \bm{\hat{Y}}_{T}) = f_{\Theta}(\bm{X}_{0}),
    \end{aligned}
\end{equation}
where $\Theta$ denotes the parameters of the neural network. 
Each prediction $\bm{\hat{Y}}_{t} \in \mathbb{R}^{H \times W}$ represents the accumulated precipitation over the preceding 24-hour window $[t-24, t]$ with $t \in \{24, 48, \cdots, T\}$. 
In this work, we consider forecasts up to 7 days ahead ($T = 168$).

\begin{figure}[!t]
    \centering
    \includegraphics[width=1\textwidth]{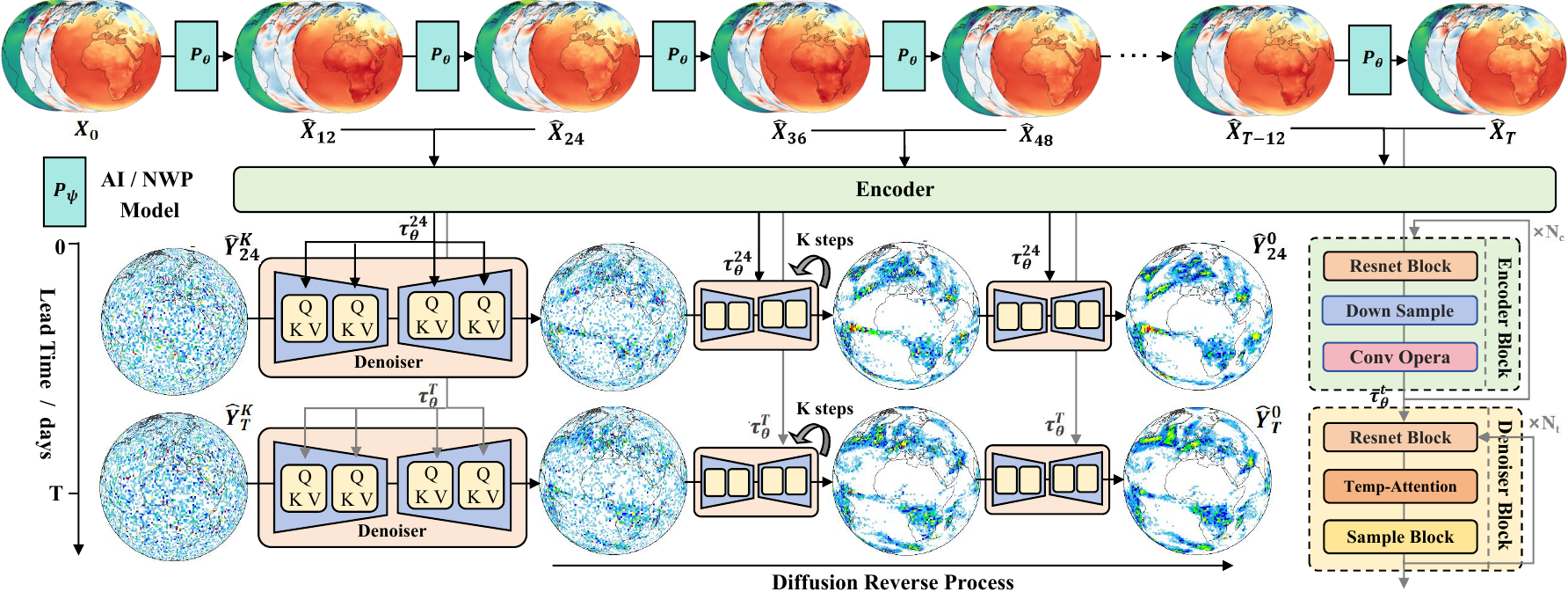} 
    \caption{Overview of our two-step framework. The workflow cycles as follows: (1) An AI/NWP model produces two 12h autoregressive forecasts$(\bm{\hat{X}}_{t-12}, \bm{\hat{X}}_t)$, which the Encoder converts into the features. (2) Guided by these features, the Denoiser generates a 24h accumulated precipitation forecast $\bm{\hat{Y}}_t$ via a $K$-step reverse diffusion process. The details are shown in the bottom right.}
    \label{fig_framework}
    
\end{figure}

\paragraph{Diffusion Model}
Diffusion models~\citep{ho2020denoising} are probabilistic frameworks designed to learn the target distribution $p(\bm{Y}_{t})$ by reversing a fixed Markov Chain of length $K$. 
In our work, the reverse process transforms Gaussian noise into a precipitation field.
Specifically, the diffusion model consists of two processes: a forward noising process and a reverse denoising process.

In the forward process, Gaussian noise is progressively injected into ground-truth precipitation fields $\bm{Y}_{t}^{0}=\bm{Y}_{t}$ and the $k$-th diffusion state $\bm{Y}^{k}_{t}$ is defined as:
\begin{equation}
    q(\bm{Y}^{k}_{t}|\bm{Y}^{0}_{t}) = \mathcal{N}(\bm{Y}^{k}_{t}; \sqrt{\overline{\alpha}_{k}} \bm{Y}^{0}_{t}, (1-\overline{\alpha}_{k})I),
\end{equation}
where $\overline{\alpha}_{k} = \prod_{j=1}^{k} \alpha_j$, $\alpha_j = 1 - \beta_j$, and $\beta_j \in (0, 1)$ is predefined by a variance schedule.
Regarding the reverse process, the target precipitation field is reconstructed via a Markov chain characterized by parameterized Gaussian transitions:
\begin{equation}
    \begin{aligned}
        p_\theta(\bm{Y}^{k-1}_{t}|\bm{Y}^{k}_{t}) = \mathcal{N}(\bm{Y}^{k-1}_{t};\mu_{\theta}(\bm{Y}^{k}_{t},k), \sigma_k^2I),
    \end{aligned}   
\end{equation}
where $\mu_\theta$ represents the posterior mean function. The reverse function $p_\theta(\bm{Y}^{k-1}_{t} | \bm{Y}^{k}_{t})$ aims to remove the noise in the precipitation field. 
While this reverse transition is probabilistic, the most successful frameworks~\citep{ho2020denoising, dhariwal2021diffusion} leverage a variant of the variational lower bound, which mirrors denoising score-matching~\citep{song2020score}. 
These models can be interpreted as a sequence of denoising methods $\epsilon_\theta(\bm{Y}^{k}_{t}, k)$, optimized to predict the noise $\epsilon \sim \mathcal{N}(0, I)$ in a degraded precipitation map $\bm{Y}_t$. 
Following the objective, the loss function is defined as:
\begin{equation}
    \mathcal{L} = \mathbb{E}_{\bm{Y}_{t}^{0}, k, \epsilon} \left\| \epsilon - \epsilon_{\theta}(\bm{Y}^{k}_{t}, k) \right\|^2,
    \label{equ:loss}
\end{equation}
where the $k$-th denoising state $\bm{Y}_{t}^{k}=\sqrt{\overline{\alpha}_k}\bm{Y}^{0}_{t} + \sqrt{1-\overline{\alpha}_k}\epsilon.$

\section{Diffusion Precipitation Generation}

\subsection{Dynamics--Precipitation Decoupling}

Instead of training a full weather prediction model from scratch, as illustrated in Figure~\ref{fig_framework}, we decouple the task into: (1) obtaining future atmospheric forecasts; (2) generating precipitation fields via conditioned diffusion. 
This design allows our framework to reuse forecast atmospheric states from existing weather models to generate precipitation aligned with a new target product (e.g., IMERG) through a learnable mapping, without expensive end-to-end retraining across different systems.

In the first step, forecasts of non-precipitation atmospheric variables (e.g., temperature, wind, and humidity) are obtained from a base weather model, which can be either an NWP system (e.g., HRES) or a WFM (e.g., GraphCast and Pangu-Weather). The base model remains frozen and only provides conditioning fields, with no gradient propagated through it.
As shown in Figure~\ref{fig:top_figure}(b), direct reuse of the ERA5-trained generator performs better than forecast-field fine-tuning, indicating that adapting to forecast states affected by accumulated lead-time errors can weaken the learned atmosphere--precipitation relationship.
We therefore use ERA5 reanalysis states $\bm{X}_{t}$ during training and atmospheric forecasts $\bm{P}_{\psi}(\bm{X}_{0}, t)$ during prediction,
where $\bm{P}_{\psi}(\cdot)$ is the base forecasting model with frozen parameters $\psi$.
$\bm{\hat{X}}_{0} = \bm{X}_{0}$ denotes the initial global weather state and $\bm{\hat{X}}_{t}$ represents the atmospheric variables at hour $t \in \{12, 24, \cdots, T-12, T\}$ in 12-hour intervals.
For each 24-hour precipitation target, the two states at $t-12$ and $t$ provide a compact description of the intraday atmospheric change. A denser sequence at 6-hour intervals is reserved for the temporal-density comparison in Appendix~\ref{appendix:temporal} for more detailed analysis.

In the second step, this framework employs a conditional diffusion model to generate the 24-hour accumulated precipitation field conditioned on $(\bm{\hat{X}}_{t-12},\bm{\hat{X}}_{t})$.
The diffusion process learns to reconstruct the precipitation distribution by gradually denoising from Gaussian noise.
Formally, for any $t \in \{24, 48, \cdots, T\}$, we denote the target precipitation field as $\bm{Y}_{t} = \bm{Y}^{0}_{t}$, and let $\bm{Y}^{1}_{t}, \cdots, \bm{Y}^{K}_{t}$ represent the latent variables in the diffusion chain.
Assuming a $K$-step denoising process, the conditional distribution factorizes as
\begin{equation}
    \begin{aligned}
        p_{\theta}(\bm{Y}^{0:K}_{t}|\bm{\hat{X}}_{t-12}, \bm{\hat{X}}_t) = p(\bm{Y}^{K}_{t}) \prod_{k=1}^{K} p_{\theta}(\bm{Y}^{k-1}_{t}|\bm{Y}^{k}_{t},\bm{\hat{X}}_{t-12}, \bm{\hat{X}}_{t}),
        \label{equ_denoising}
    \end{aligned}
\end{equation}
where $\bm{Y}^{K}_{t} \sim \mathcal{N}(0, I)$. 
This formulation estimates the conditional precipitation distribution at each forecast horizon under its corresponding atmospheric condition.
In this formulation, the denoising process is no longer a blind reconstruction of noise but a constrained synthesis guided by the evolution of atmospheric states. 
We detail the framework design of the Encoder and the Conditional Denoiser that realizes this complex mapping $f_{\Theta}$ in the following section.

\paragraph{Training}
Based on the atmosphere-conditioning pairs, we then learn the diffusion precipitation generation via
\begin{equation}
    \mathcal{L} = \mathbb{E}_{\bm{Y}_{t}^{0}, \bm{\hat{X}}, k, \epsilon} \left\| \epsilon - \epsilon_{\theta}(\bm{Y}^{k}_{t}, k, \tau_{\theta}(\bm{\hat{X}}_{t-12}, \bm{\hat{X}}_{t})) \right\|^2,
    \label{equ:final_loss}
\end{equation}
where $k$ is sampled from $\{1,\ldots,K\}$, $\epsilon \sim \mathcal{N}(0, I)$, and both $\tau_{\theta}$ and $\epsilon_{\theta}$ are optimized via Eq.~\eqref{equ:final_loss}.

\subsection{Precipitation Generation Architecture}
As mentioned earlier, we integrate forecasts $\bm{\hat{X}}_t$ from NWP systems or WFMs $\bm{P}_\psi$ into the diffusion framework as conditioning information for generating the corresponding target precipitation field.
Drawing inspiration from generative models~\citep{rombach2022high}, diffusion models (DMs) are in principle capable of modeling conditional distributions of the form $p(\bm{Y}_{t}|\bm{\hat{X}}_{t-12},\bm{\hat{X}}_{t})$.
This can be implemented with a conditional denoising method $\epsilon_{\theta}(\bm{Y}^{k}_{t}, k, \tau_{\theta}(\bm{\hat{X}}_{t-12}, \bm{\hat{X}}_{t}))$.
\textbf{However}, directly feeding raw multivariate atmospheric fields into the denoiser is memory-intensive due to their high dimensionality. Furthermore, sequential atmospheric states contain substantial spatiotemporal redundancy. To address this, we use a convolutional encoder to compress multi-level forcing into a latent representation, reducing computation while preserving precipitation-relevant signals.

\paragraph{\textbf{Multi-Scale Atmospheric Encoder}}
Following the previous studies~\citep{trenberth2003changing, bi2023accurate, espeholt2022deep}, we propose an \textbf{Atmospheric Encoder} to effectively transform atmospheric forecasts $(\bm{\hat{X}}_{t-12}, \bm{\hat{X}}_t)$ into an intermediate latent representation:
$\tau_{\theta}^t=\tau_{\theta}(\bm{\hat{X}}_{t-12}, \bm{\hat{X}}_t)$. 
As shown in Figure~\ref{fig_framework}, each Encoder block consists of a ResNet block, a sampling operator, and a convolutional operator. First, let $l \in \{1, \dots, L\}$ denote the index of the downsampling stage. The feature extraction at layer $l$ is formulated as:
\begin{align}
 \bm{R}^{l}_t = \operatorname{Conv}\!\left(
 \operatorname{\sigma}(\operatorname{GroupNorm}(\bm{F}^{l-1}_t))\right),\ \bm{F}^{l}_t = \bm{F}^{l-1}_t+\bm{R}^{l}_t,\ \tilde{\bm{F}}^{l}_t = \operatorname{Downsample}(\bm{F}^{l}_t)
\end{align}
Here, $\sigma$ denotes the SiLU function, and $\mathbf{F}^{l}_t$ is passed to the next level after spatial downsampling. In the initial compression stage, this operation maps a feature tensor from $C\times H\times W$ to $4C\times(H/4)\times(W/4)$, reducing its feature volume to one quarter while increasing the channel capacity.

Unlike standard CNNs that treat inputs as static channels, our encoder integrates a ConvGRU-based gating mechanism within each hierarchical stage to model the temporal evolution of atmospheric states.
Specifically, we employ a Convolutional GRU (ConvGRU) cell as the "convolutional operator". Let $s\in\{1,2\}$ index the states at $t-12$ and $t$, respectively, and let $\tilde{\mathbf{F}}_s^l$ denote the corresponding downsampled feature at scale $l$. We initialize $\mathbf{H}^{l}_{0}=0$, and the ConvGRU updates its hidden state as:
\begin{align}
    \mathbf{r}^{l}_s &= \sigma_g(\mathbf{W}^{l}_r * [\tilde{\mathbf{F}}^{l}_s,\mathbf{H}^{l}_{s-1}]+\mathbf{b}^{l}_r), \quad
    \mathbf{z}^{l}_s = \sigma_g(\mathbf{W}^{l}_z * [\tilde{\mathbf{F}}^{l}_s,\mathbf{H}^{l}_{s-1}]+\mathbf{b}^{l}_z), \\
    \tilde{\mathbf{H}}^{l}_s &= \operatorname{tanh}(\mathbf{W}^{l}_h * [\tilde{\mathbf{F}}^{l}_s, \mathbf{r}^{l}_s \odot \mathbf{H}^{l}_{s-1}]+\mathbf{b}^{l}_h), \quad
    \mathbf{H}^{l}_s = (1 - \mathbf{z}^{l}_s) \odot \mathbf{H}^{l}_{s-1} + \mathbf{z}^{l}_s \odot \tilde{\mathbf{H}}^{l}_s.
\end{align}
Here, $*$ denotes convolution, $\odot$ is the Hadamard product, and $\sigma_g$ is the sigmoid function. After processing the two states, the final hidden state $\mathbf{H}^l_2$ encodes their temporal change at scale $l$. The resulting multi-scale features form the condition $\tau_{\theta}^t$:
\begin{equation}
\tau_{\theta}^t = \operatorname{Embedding}(\mathbf{H}^1_2, \mathbf{H}^2_2, \dots, \mathbf{H}^L_2)
 = \{\tau_{\theta,i}^t\}_{i=1}^{L}.
\end{equation}
Each feature $\tau_{\theta,i}^t\in\mathbb{R}^{M_i\times d_c}$ is projected into the denoising block at the matching U-Net resolution. These features capture scale-dependent temporal variations more efficiently than the raw fields $\bm{X}_t$.

\paragraph{\textbf{Denoiser with Attention mechanism}}
Daily precipitation reflects accumulated effects of atmospheric processes acting across time and space~\citep{trenberth2003changing}. This nonlocal correspondence motivates spatial alignment between atmospheric conditions and daily precipitation features. We therefore introduce specialized conditioning to \textbf{maintain spatial consistency during denoising} and \textbf{bridge this cross-field gap}. We augment the U-Net backbone with a cross-attention mechanism~\citep{vaswani2017attention,jaegle2021perceiver1,jaegle2021perceiver2}. At the $i$-th U-Net resolution, let $\bm{Z}_i=\varphi_i(\bm{Y}_{t}^{k})\in\mathbb{R}^{N_i\times d_i}$ denote the flattened denoising feature and $\bm{C}_i=\tau_{\theta,i}^t\in\mathbb{R}^{M_i\times d_c}$ the corresponding atmospheric condition. Cross-attention is implemented as
\[
\operatorname{Attention}(Q,K,V)=\operatorname{softmax}(QK^T/\sqrt{d_a})V,
\]
with
\begin{equation}
    Q = \bm{Z}_i W_{Q}^{(i)}, \qquad
    K = \bm{C}_i W_{K}^{(i)}, \qquad
    V = \bm{C}_i W_{V}^{(i)}.
\end{equation}
Here, $d_a$ is the attention dimension, $W_{Q}^{(i)}\in\mathbb{R}^{d_i\times d_a}$, and $W_{K}^{(i)},W_{V}^{(i)}\in\mathbb{R}^{d_c\times d_a}$ are learnable channel projections. This interaction allows precipitation features to selectively attend to atmospheric conditions at the matching spatial scale. Based on the above analysis, the Denoiser can predict the noise $\epsilon$ on the precipitation field. With Cross-attention, the prediction function can be expressed as:
\begin{equation}
\hat{\epsilon}_\theta = \text{Denoiser}(\bm{Y}_{t}^{k}, \tau_\theta^{t}, k) = f_{out}  \left( \text{Cross-Attn}(\bm{Y}_{t}^{k}, \tau_\theta^{t}) \right)
\end{equation}
The output function $f_{out}$ restores the feature maps to the original resolution via the Sample Block and utilizes a convolutional layer to produce the predicted noise $\hat{\epsilon}_\theta$ with dimensions identical to $\bm{Y}_t^k$.

\section{Experiments}

\subsection{Experimental Setup}

\paragraph{Datasets}
We construct a domain-specific dataset integrating atmospheric analyses, observation-based precipitation, heterogeneous forecasts, and station measurements. Hourly \textbf{ERA5} states~\citep{hersbach2020era5}, using the variables and levels in Table~\ref{tab:data}, are paired by valid time with IMERG Final Run V07B 24-hour precipitation over $t-24,t$~\citep{pradhan2022review}. HRES and ENS are sourced from WeatherBench2~\citep{rasp2024weatherbench}, while GraphCast and Pangu-Weather forecasts are generated with their released models. All products are aligned to $0.25^\circ$ and $1.5^\circ$ grids. Data span 1998--2023, with 1998--2021 for training, 2023 for validation, and 2022 for testing. The 2022 test year matches WeatherBench2 forecast availability. Finally, 2,288 China Meteorological Administration stations are reserved for external evaluation of localized heavy-rainfall events.

\paragraph{Metrics}
Following \citep{bi2023accurate, gao2023prediff, sonderby2020metnet}, we quantify precipitation accuracy using the Anomaly Correlation Coefficient (\textbf{ACC}) and Root Mean Square Error (\textbf{RMSE}), and event skill using the Critical Success Index (\textbf{CSI}) and Heidke Skill Score (\textbf{HSS}). Probabilistic skill is evaluated with the Continuous Ranked Probability Score (\textbf{CRPS}) and Brier Score (\textbf{BS}). Deterministic results use one generated sample per case without sample selection. Metric definitions are provided in Section~\ref{app:metrics} of the Appendix.

\begin{figure}[t]
    \centering
    
    \includegraphics[width=1\textwidth]{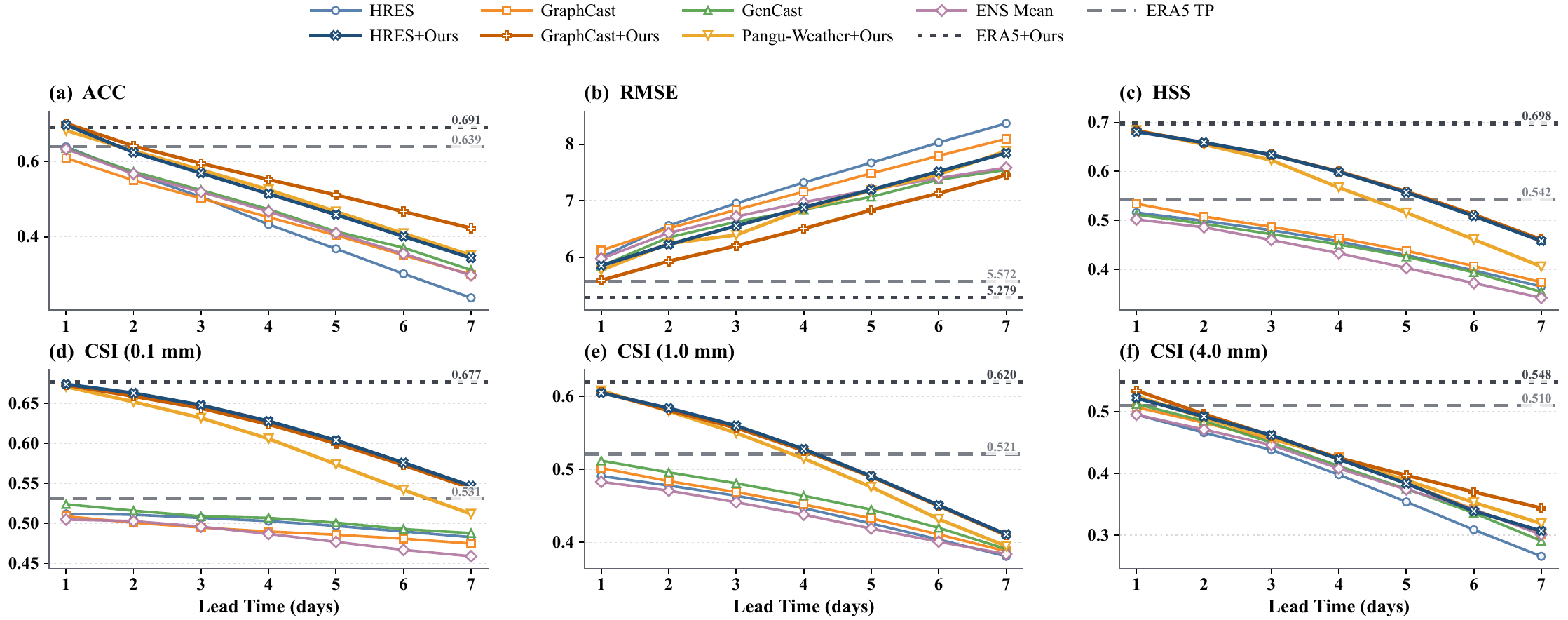} 
    
    \caption{Global $1.5^\circ$ deterministic skill over the 2022 test set for forecast backbones, their variants with our framework, and external system references. IMERG is used as the verification reference.}
    \label{fig:metrics}
    
\end{figure}

\paragraph{Baselines}
We evaluate our framework with the data-driven \textbf{Pangu-Weather}~\citep{bi2023accurate} and \textbf{GraphCast}~\citep{lam2023learning} backbones, as well as the numerical \textbf{HRES} system~\citep{haiden2021evaluation}. Pangu-Weather provides no native precipitation output, while GraphCast and HRES enable direct comparison with their own precipitation forecasts. For probabilistic evaluation, we additionally include \textbf{GenCast}~\citep{price2025probabilistic} and \textbf{ENS}~\citep{leutbecher2017stochastic} as external system-level baselines, since their ensembles also capture uncertainty in atmospheric trajectories.

\begin{table}[tbp]
  
  \centering
  \captionsetup{font=footnotesize,skip=3pt}
  \caption{Probabilistic forecasting for a 50-member ensemble.}
  \label{tab:pro_metrics}

  \setlength{\tabcolsep}{2pt}
  \renewcommand{\arraystretch}{1.1}
  \small
    \begin{tabular*}{\linewidth}{@{\extracolsep{\fill}}llccccccc@{}}
    \toprule
    \multirow{2.5}{*}{\textbf{Metrics}} & \multirow{2.5}{*}{\textbf{Models}} & \multicolumn{7}{c}{\textbf{Lead Time (Days)}} \\
    \cmidrule(lr){3-9}
    & & \textbf{1} & \textbf{2} & \textbf{3} & \textbf{4} & \textbf{5} & \textbf{6} & \textbf{7} \\
    \midrule
    \multirow{5}{*}{\text{CRPS} $\downarrow$} 
    & ENS                  & 1.528 & 1.740 & 1.824 & 1.975 & 2.161 & 2.346 & 2.484 \\
    & GenCast              & 1.512 & 1.651 & 1.778 & 1.943 & 2.147 & 2.255 & 2.368 \\
    & Pangu+\textbf{Ours}     & \underline{1.483} & \underline{1.619} & 1.756 & 1.910 & 2.061 & 2.208 & 2.336 \\
    & GraphCast+\textbf{Ours} & \textbf{1.477} & \textbf{1.569} & \textbf{1.660} & \textbf{1.769} & \textbf{1.898} & \textbf{2.035} & \textbf{2.180} \\
    & HRES+\textbf{Ours}      & 1.525 & 1.637 & \underline{1.747} & \underline{1.867} & \underline{2.006} & \underline{2.156} & \underline{2.302} \\
    \midrule
    \multirow{5}{*}{\text{BS} $\downarrow$} 
    & ENS                  & 0.193 & 0.201 & 0.211 & 0.229 & 0.237 & 0.239 & 0.245 \\
    & GenCast              & 0.177 & 0.195 & 0.209 & 0.216 & 0.230 & 0.240 & 0.247 \\
    & Pangu+\textbf{Ours}     & 0.141 & 0.149 & 0.164 & 0.180 & 0.200 & 0.221 & 0.240 \\
    & GraphCast+\textbf{Ours} & \textbf{0.138} & \underline{0.146} & \underline{0.155} & \textbf{0.164} & \textbf{0.181} & \textbf{0.199} & \textbf{0.218} \\
    & HRES+\textbf{Ours}      & \underline{0.140} & \textbf{0.145} & \textbf{0.154} & \underline{0.166} & \underline{0.185} & \underline{0.203} & \underline{0.219} \\
    \bottomrule
  \end{tabular*}
  
\end{table}

\paragraph{Implementation} 
Our framework is implemented in PyTorch \citep{paszke2019pytorch} and trained for 300K steps. We use the AdamW optimizer \citep{kingma2014adam,loshchilov2017decoupled} with a learning rate of $1\times10^{-5}$ and $(\beta_1,\beta_2)=(0.9,0.99)$. Following standard diffusion settings \citep{rombach2022high}, we use $1000$ diffusion steps. During inference, DDIM \citep{songdenoising} reduces the denoising process to $100$ steps. All experiments are conducted on four NVIDIA A800 GPUs. At both $0.25^\circ$ and $1.5^\circ$ resolutions, the resolution-specific checkpoint is applied to forecast states from all aligned backbones without backbone-specific retraining. For probabilistic evaluation, we generate $M=50$ members from independently sampled Gaussian noise. The forecast-field fine-tuning diagnostic is described in Appendix~\ref{app:cross_backbone}.

\subsection{Global Assessment at \texorpdfstring{$1.5^\circ$}{1.5 degrees}}

\paragraph{Deterministic Forecast}
We evaluate Pangu-Weather, GraphCast, and HRES equipped with our framework for lead times up to 7 days in Figure~\ref{fig:metrics}. Following \citet{andrychowicz2023deep}, CSI thresholds are set to $\{0.1, 1, 4\}$ mm. Based on these results, we have the following findings: 
\begin{itemize}[leftmargin=*]
    
    \item Our framework enables Pangu-Weather to forecast precipitation and raises GraphCast's ACC by 0.103 while reducing its RMSE by 0.623 on average. GraphCast+Ours also outperforms GenCast. HRES+Ours performs comparably to GraphCast+Ours and better than Pangu-Weather+Ours, confirming that precipitation generation still depends on backbone atmospheric skill.
    \item ERA5+Ours serves as an analysis-conditioned diagnostic: its improvement over ERA5 total precipitation reflects closer agreement with IMERG, rather than an operational forecasting gain.
    \item HSS and CSI show consistent gains in precipitation-event discrimination across the thresholds. Improvements at 0.1 and 1.0 mm indicate better detection of small daily accumulations, while the 4-mm threshold supports improved detection of larger amounts within the evaluated range.
\end{itemize}

\paragraph{Probabilistic Forecast}
To investigate ensemble precipitation skill, we compare 50-member forecasts from our framework with \textbf{ENS} and \textbf{GenCast} in Table~\ref{tab:pro_metrics}. From the results, we can find that:
\begin{itemize}[leftmargin=*]
    
    \item Overall, our variants are competitive with the external ensembles and achieve lower CRPS in most comparisons. GraphCast+Ours performs best at every reported lead time. At Day 7, its CRPS is 2.180, compared with 2.484 for ENS and 2.368 for GenCast. The lower CRPS indicates closer agreement between the predicted precipitation distributions and IMERG.
    \item All our variants achieve lower BS than ENS and GenCast at every lead time, indicating more accurate precipitation-event probabilities. At Day 7, GraphCast+Ours achieves 0.218, compared with 0.245 for ENS and 0.247 for GenCast, demonstrating stronger probabilistic event forecasting.
\end{itemize}

\begin{figure}[t]
    \centering
    
    \includegraphics[width=1\textwidth]{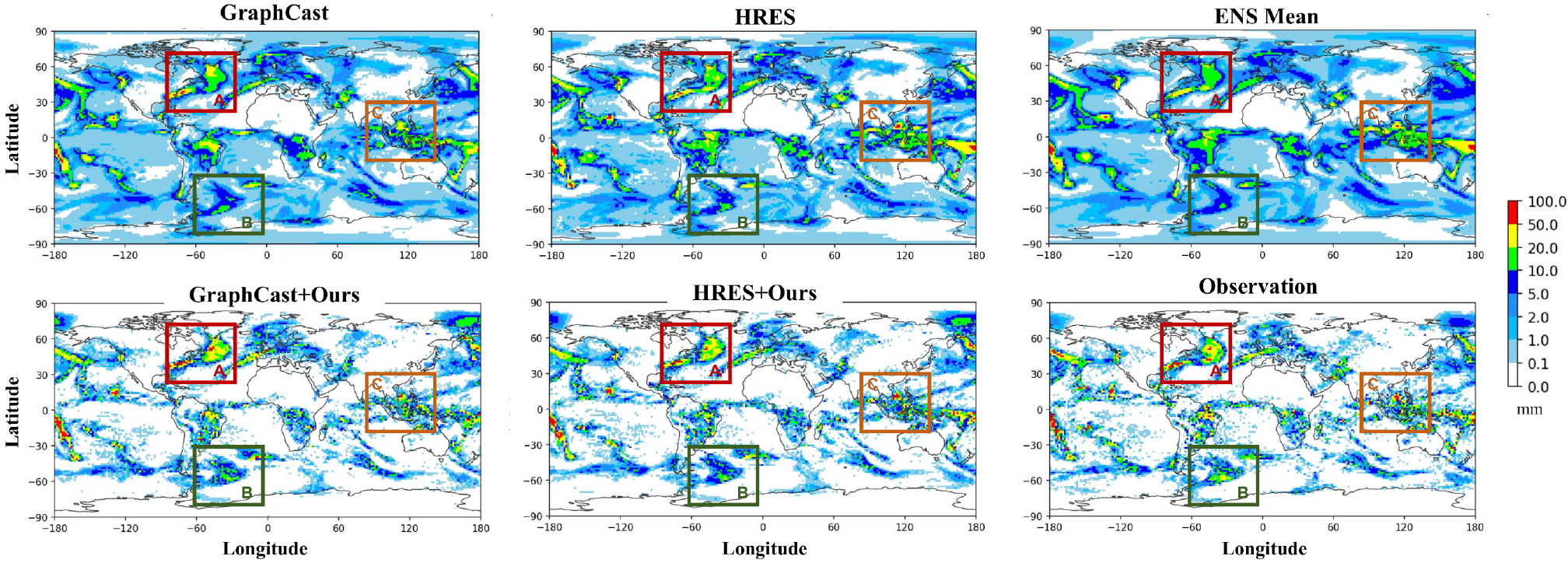}
    
    \caption{Spatial comparison of Day-3 precipitation forecasts from the baseline models, their variants with our framework, and the IMERG reference. Boxes A--C highlight regional differences.}
    \label{fig:visual}

\end{figure}

\begin{figure}[tbp]
    
    \centering
    \includegraphics[width=1\textwidth]{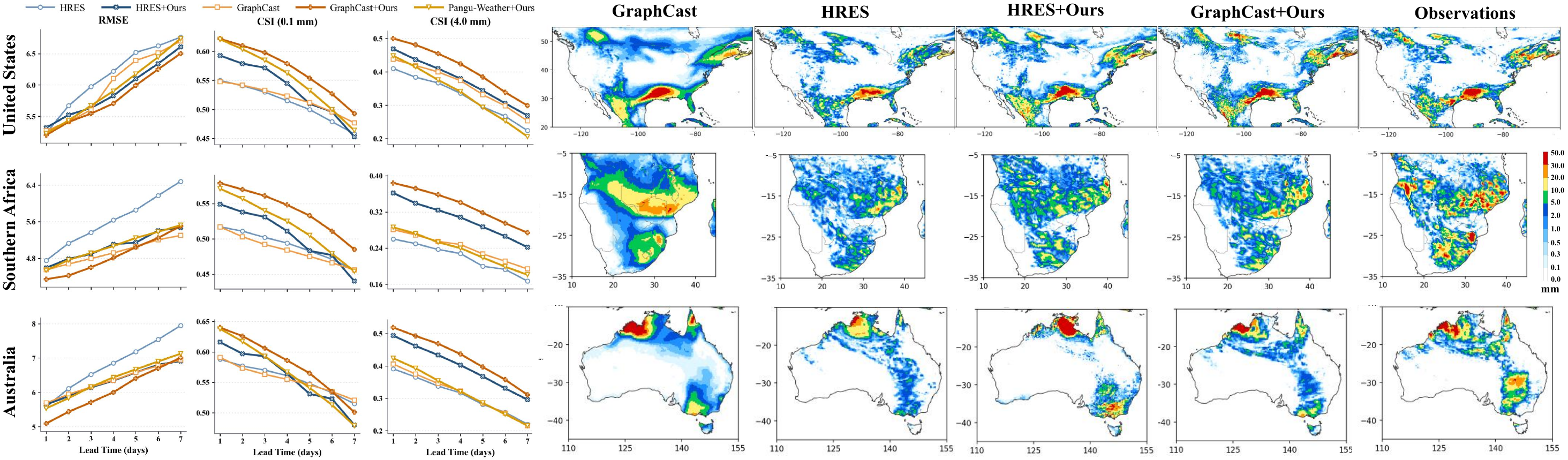}
    
    \caption{Regional evaluation over the United States, Southern Africa, and Australia. Left: RMSE and CSI across lead times. Right: precipitation fields from WFMs, their variants, and Observations.}
    \label{fig:region_comparison}

\end{figure}

\begin{figure}[tbp]

    \centering
    \includegraphics[width=\columnwidth]{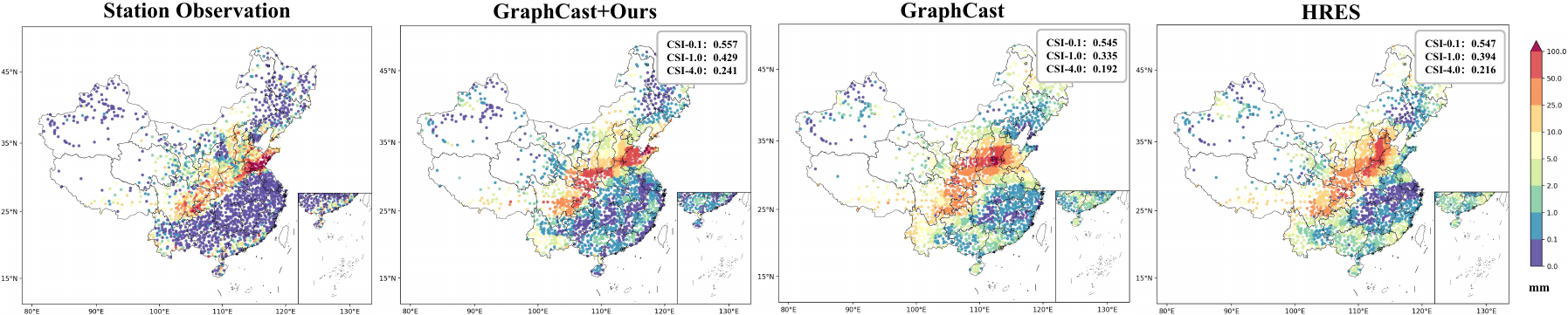} 
    
    \caption{Station-based evaluation over North China in July 2022, initialized on June 29.}
    \label{fig:station_map}

\end{figure}

\paragraph{Visualization}
We visualize Day-3 precipitation in Figure~\ref{fig:visual}. All systems capture the large-scale patterns, while the base forecasts are smoother at local scales. As expected, averaging ENS members further suppresses member-level variability.
With our framework, forecasts show more distinct local accumulation maxima. In Boxes A and C, their spatial patterns and daily amounts better match IMERG, while Box B shows clearer rainband boundaries. These examples support improved spatial representation of daily precipitation, although visual sharpness alone does not establish forecast accuracy.
More visualizations are shown in Figure~\ref{fig:all_visual} of the Appendix.

\subsection{Regional Evaluation at \texorpdfstring{$0.25^\circ$}{0.25 degrees}}

\paragraph{Region Comparisons.} 
Figure~\ref{fig:region_comparison}(left) evaluates our framework over the United States, Southern Africa, and Australia at $0.25^\circ$. Across these regions, both HRES+Ours and GraphCast+Ours improve precipitation forecasting, with GraphCast+Ours achieving the best overall performance. These results demonstrate robust coupling with aligned $0.25^\circ$ WFM and NWP outputs and effective use of their fine-scale atmospheric structures for regional precipitation prediction. Figure~\ref{fig:region_comparison}(right) compares 24-h precipitation forecasts. GraphCast oversmooths rainfall, while HRES misses local maxima despite capturing large-scale patterns. Our framework preserves this organization and recovers sharper, heterogeneous daily rainfall patterns over Southern Africa and northern Australia. 
Closer agreement with IMERG further indicates that our framework better represents localized daily accumulation patterns at $0.25^\circ$.

\begin{figure}[t] %
    \centering
    
    \includegraphics[width=1\columnwidth]{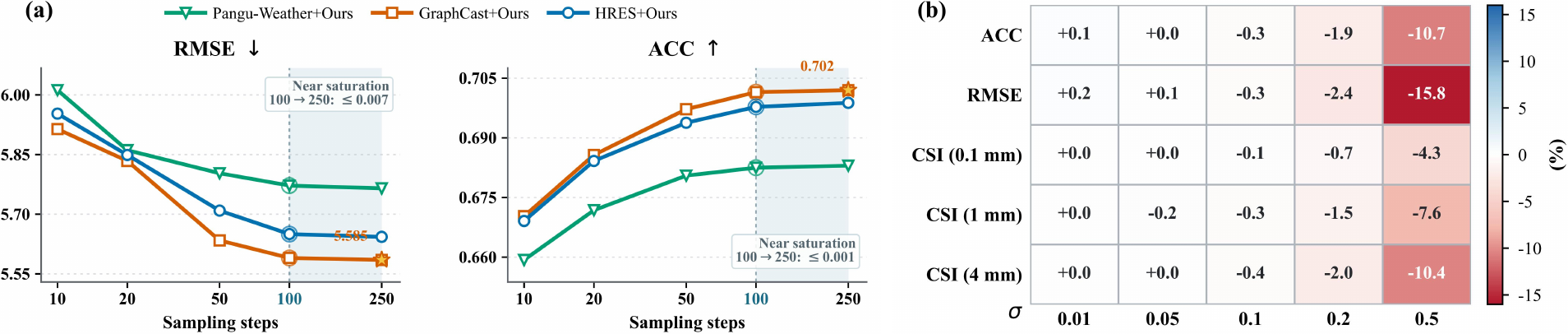} 
    
    \caption{(a) Impact of sampling steps. RMSE and ACC approach saturation at 100 steps. (b) Relative performance changes under Gaussian perturbed inputs; red indicates degradation.}
    \label{fig:steps&noise_robustness}

\end{figure}

\paragraph{Case Study at 2,288 Stations}
We examine North China's early-July 2022 heavy-rainfall event, where Chaba's remnants and a westerly trough organized moisture transport and a broad rainband with local maxima. In Figure~\ref{fig:station_map}, HRES and GraphCast capture the rainband but smooth contrasts in daily totals. GraphCast+Ours preserves its placement and recovers accumulation maxima without domain-wide intensification. Multi-threshold CSI gains support improved event detection.

\subsection{Mechanism and Robustness Analysis}
\label{sec:robustness_analysis}

\begin{table}[tbp]
  
  \centering
  \captionsetup{font=footnotesize,skip=3pt}
  \caption{Day-1 configuration comparison. ``--'' denotes an omitted proposed module; U-Net and CorrDiff replace the complete generator.}
  \label{tab:ablation_study}

  \setlength{\tabcolsep}{2pt}
  \renewcommand{\arraystretch}{1.1}
  \small
  \begin{tabular*}{\linewidth}{@{\extracolsep{\fill}}lccccccccc@{}}
    \toprule
    \textbf{Condition} & \textbf{Temp} & \textbf{Align} & \textbf{Generator}
    & \textbf{ACC$\uparrow$} & \textbf{RMSE$\downarrow$} & \textbf{HSS$\uparrow$}
    & \textbf{CSI$_{0.1}\uparrow$} & \textbf{CSI$_1\uparrow$} & \textbf{CSI$_4\uparrow$} \\
    \midrule
    \multirow{8}{*}{GraphCast}
    & -- & -- & Native & 0.6091 & 6.120 & 0.5347 & 0.5097 & 0.4928 & 0.4271 \\
    & GRU & -- & Diff. & 0.6217 & 5.924 & 0.5914 & 0.5612 & 0.5267 & 0.4826 \\
    & LSTM & -- & Diff. & 0.5942 & 6.318 & 0.5165 & 0.4942 & 0.4725 & 0.3945 \\
    & -- & Attn. & Diff. & \underline{0.6571} & \underline{5.721} & 0.6213 & 0.6158 & \underline{0.5572} & \underline{0.5185} \\
    & -- & Gated & Diff. & 0.6126 & 6.067 & 0.5678 & 0.5367 & 0.5126 & 0.4438 \\
    \cmidrule(lr){2-10}
    & -- & -- & U-Net & 0.4071 & 9.840 & \underline{0.6268} & \underline{0.6458} & 0.5464 & 0.4535 \\
    & -- & -- & CorrDiff & 0.5004 & 6.939 & 0.5733 & 0.5877 & 0.5067 & 0.4263 \\
    & GRU & Attn. & Diff. (Ours) & \textbf{0.7015} & \textbf{5.590} & \textbf{0.6832} & \textbf{0.6736} & \textbf{0.6061} & \textbf{0.5340} \\
    \midrule
    \multirow{8}{*}{HRES}
    & -- & -- & Native & 0.6394 & 6.010 & 0.5164 & 0.5126 & 0.4913 & 0.4151 \\
    & GRU & -- & Diff. & 0.6385 & 5.984 & 0.5927 & 0.5581 & 0.5357 & 0.4782 \\
    & LSTM & -- & Diff. & 0.5905 & 6.359 & 0.5118 & 0.4909 & 0.4769 & 0.3916 \\
    & -- & Attn. & Diff. & \underline{0.6572} & \underline{5.841} & 0.6174 & 0.6087 & \underline{0.5581} & \underline{0.5108} \\
    & -- & Gated & Diff. & 0.6083 & 6.132 & 0.5541 & 0.5318 & 0.5118 & 0.4397 \\
    \cmidrule(lr){2-10}
    & -- & -- & U-Net & 0.4023 & 9.733 & \underline{0.6267} & \underline{0.6451} & 0.5479 & 0.4521 \\
    & -- & -- & CorrDiff & 0.4911 & 7.095 & 0.5719 & 0.5888 & 0.5064 & 0.4233 \\
    & GRU & Attn. & Diff. (Ours) & \textbf{0.6978} & \textbf{5.650} & \textbf{0.6818} & \textbf{0.6742} & \textbf{0.6054} & \textbf{0.5221} \\
    \bottomrule
    \end{tabular*}
  
\end{table}

\paragraph{Effectiveness of the Model Design}
Table~\ref{tab:ablation_study} compares partial and complete configurations under the same data split and evaluation protocol. \textbf{(i) Temporal extraction.} Among the temporal configurations, ConvGRU consistently outperforms ConvLSTM under both conditioning backbones, supporting its use for compact atmospheric evolution encoding. \textbf{(ii) Spatial alignment.} Among the alignment configurations, cross-attention performs better than gated fusion, showing the benefit of selectively aligning atmospheric and precipitation features. \textbf{(iii) Complete-model comparison.} U-Net remains competitive for low-intensity event detection but is less accurate for the overall intensity distribution, while CorrDiff's lower scores are consistent with a mismatch between its coarse-to-fine objective and this same-resolution generation setting. The full configuration performs best across the reported metrics, indicating complementary gains from temporal extraction and spatial alignment.

\paragraph{Efficiency and Robustness Analysis}
As shown in Figure~\ref{fig:steps&noise_robustness}(a), increasing the sampling steps from 10 to 100 consistently improves RMSE and ACC across all three backbones. Beyond 100 steps, the performance curves largely saturate, with only marginal gains up to 250 steps. We therefore use 100 sampling steps by default, providing a favorable trade-off between forecast quality and computational cost. In addition, we perturb normalized atmospheric inputs with Gaussian noise of standard deviation $\sigma$. As shown in Figure~\ref{fig:steps&noise_robustness}(b), performance changes remain limited for small perturbations and degrade gradually as $\sigma$ increases, indicating controlled sensitivity to synthetic input errors.

\section{Conclusion}

We present \emph{Dynamics--Precipitation Decoupling} for daily precipitation generation at medium-range lead times. Trained on ERA5--IMERG pairs, the diffusion generator maps non-precipitation atmospheric states to daily accumulations through recurrent encoding and cross-field alignment. It does not use the backbone's precipitation outputs. One generator per resolution is reused across aligned backbones without backbone-specific retraining. Global and regional evaluations show improved deterministic skill, with competitive probabilistic performance in the global evaluation. A heavy-rainfall case study at 2,288 stations further supports improved daily accumulation patterns and event detection. Ablations support complementary contributions from temporal encoding and spatial alignment. The framework samples precipitation conditional on a fixed atmospheric trajectory, and its skill remains dependent on the backbone forecast. Further evaluation across forecasting systems and heavy-rainfall events remains important. These results demonstrate the potential of decoupled precipitation generation as a reusable component of medium-range forecasting systems.

\clearpage
\appendix
\subsection*{Ethics statement and limitations}
Cross-backbone reuse is empirically validated only for the aligned forecast systems considered in this work, while the $1.5^\circ$ and $0.25^\circ$ settings require separately trained models. Training on ERA5 states also introduces a distribution shift when conditioning on imperfect forecasts, which we probe through noise perturbation and forecast-field fine-tuning experiments. Future work should further evaluate transfer across other reanalyses, such as JRA and MERRA-2, and operational products such as GFS. IMERG retrieval uncertainty and the additional cost of diffusion sampling remain further limitations. All datasets are publicly available and contain no personal information; nevertheless, generated precipitation should be used as decision support and requires regional validation, expert oversight, and clear communication of forecast uncertainty.

\subsection*{AI use statement}
In this work, we used LLMs as an auxiliary tool to polish the writing and improve the manuscript's clarity. Specifically, we use it to review the entire text for grammatical accuracy, improve sentence structure, and ensure consistent phrasing and tone throughout the paper. 

\section{Appendix}

\subsection{Baselines}

\begin{itemize}
\item \textbf{Pangu-Weather} Developed by Huawei Cloud, Pangu-Weather is the first AI model to surpass traditional NWP accuracy. It uses four 3D Earth-specific Transformer models for 1-, 3-, 6-, and 24-hour forecasts, each with approximately 64 million parameters (256 million in total), and combines them through hierarchical temporal aggregation. The model maintains global physical consistency while delivering massive speedups in inference compared to traditional numerical systems.
\item \textbf{GraphCast} Developed by Google DeepMind, GraphCast is a mid-range forecasting model based on Graph Neural Networks (GNNs). It employs a "Grid-to-Mesh-to-Grid" architecture, mapping atmospheric variables onto a multi-scale icosahedral mesh to capture long-range dependencies and resolve polar singularities. At $0.25^\circ$ resolution, it delivers SOTA deterministic accuracy across thousands of meteorological variables.
\item \textbf{GenCast} As an advanced global deep learning forecasting model, GenCast can generate probabilistic forecasts that surpass those of traditional ensemble systems. It demonstrates superior performance in providing high-resolution, physically consistent precipitation distribution and extreme weather warnings by efficiently capturing the complex uncertainties of atmospheric evolution.
\item \textbf{HRES} Operated by ECMWF, HRES is the world's leading Numerical Weather Prediction (NWP) system and the industry's "gold standard". It describes atmospheric evolution through complex fluid dynamics and physical parameterizations, integrated with massive global observations. Despite the rise of AI models, HRES remains the essential benchmark for validating physical reliability in weather forecasting.
\item \textbf{ENS} The Ensemble Prediction System (ENS) operated by ECMWF is designed to quantify atmospheric uncertainty by generating multiple forecast members through subtle perturbations in initial conditions or model parameters. Unlike the single deterministic HRES, ENS enhances medium-range robustness via the ensemble mean (ENS-Mean) or probabilistic distributions. In precipitation forecasting, it serves as a critical operational benchmark for evaluating AI models' proficiency in uncertainty quantification and probabilistic accuracy.
\end{itemize}

\begin{table}[htbp]
    \centering
    \small
    \caption{ERA5 variable--level interface used in the experiments.}
    \label{tab:data}
    \setlength{\tabcolsep}{4pt}
    \begin{tabular}{llll}
    \toprule
    Type & Variable & Abbreviation & Levels \\
    \midrule
    Static & Land--sea mask & LSM & -- \\
    Static & Orography & ORG & -- \\
    Surface & 2-m temperature & T2m & -- \\
    Surface & 10-m zonal wind & U10 & -- \\
    Surface & 10-m meridional wind & V10 & -- \\
    Surface & Mean sea-level pressure & MSLP & -- \\
    \midrule
    Upper & Geopotential & Z & 50, 100, 150 \\
    Upper & U component of wind & U & 200, 250, 300 \\
    Upper & V component of wind & V & 400, 500, 600 \\
    Upper & Temperature & T & 700, 850 \\
    Upper & Specific humidity & Q & 925, 1000 \\
    \bottomrule
    \end{tabular}
\end{table}

\subsection{Metrics}
\label{app:metrics}

\paragraph{Root mean square error (RMSE)}
Root Mean Square Error (RMSE) evaluates the average size of the errors in a dataset. It measures how well the predicted values align with the true values, where a smaller RMSE indicates better prediction accuracy.
\begin{equation}
    \mathrm{RMSE}=\frac{1}{T}\sum_{t=1}^{T}\sqrt{\frac{1}{H\times W}\sum_{h=1}^{H}\sum_{w=1}^{W}L(h)(\hat{\bm{Y}}_{t,h,w}-\bm{Y}_{t,h,w})^{2}}
\end{equation}
where $L(h)$ is introduced as a latitude-based weighting factor:
\begin{equation}
    \label{equ:Li}
    L(h) = \frac{\cos(\operatorname{lat}(h))}{\frac{1}{H} \sum_{h=1}^{H} \cos(\operatorname{lat}(h))}
\end{equation}

\paragraph{Anomaly correlation coefficient (ACC)}
The Anomaly Correlation Coefficient (ACC) assesses forecast accuracy by quantifying the correlation between predicted and observed anomalies. It is widely used in climate and weather prediction to assess how effectively the model identifies deviations from the climatological mean. A higher ACC value signifies better forecasting performance.
\begin{equation}
    \mathrm{ACC}=\frac{\sum_{t,h,w}L(h)(\hat{\bm{Y}}_{t,h,w}-C_{t,h,w})(\bm{Y}_{t,h,w}-C_{t,h,w})}{\sqrt{\sum_{t,h,w}L(h)(\hat{\bm{Y}}_{t,h,w}-C_{t,h,w})^{2}\sum_{t,h,w}L(h)(\bm{Y}_{t,h,w}-C_{t,h,w})^{2}}}
\end{equation}
where climatology $C$ is computed only from the 1998--2021 training period. The calculation of \( L(h) \) follows Eq.~\ref{equ:Li}.

\paragraph{Critical Success Index (CSI)}
CSI measures the overlap between observed and predicted precipitation events and ranges from 0 (worst) to 1 (best). For global evaluation, contingency counts are pooled over test cases and grid cells using latitude weights.
\begin{equation}
    \mathrm{CSI} = \frac{TP}{TP + FP + FN}
\end{equation}
where the weighted contingency counts are defined as:
\begin{align}
    & TP = \sum_{t,h,w} L(h)\,\mathbb{I}[(\hat{\bm{Y}}_{t,h,w} \geq th)\land(\bm{Y}_{t,h,w} \geq th)], \\
    & FP = \sum_{t,h,w} L(h)\,\mathbb{I}[(\hat{\bm{Y}}_{t,h,w} \geq th)\land(\bm{Y}_{t,h,w} < th)], \\
    & FN = \sum_{t,h,w} L(h)\,\mathbb{I}[(\hat{\bm{Y}}_{t,h,w} < th)\land(\bm{Y}_{t,h,w} \geq th)], \\
    & TN = \sum_{t,h,w} L(h)\,\mathbb{I}[(\hat{\bm{Y}}_{t,h,w} < th)\land(\bm{Y}_{t,h,w} < th)]
    \label{equ:tpfpfn}
\end{align}
Here, $\mathbb{I}(\cdot)$ is an indicator and $th\in\{0.1,1.0,4.0\}$ mm is the precipitation threshold.

\paragraph{Heidke Skill Score (HSS)}
\label{appendix:hss}
HSS measures the accuracy of a binary forecast relative to random chance. It accounts for true positives, true negatives, false positives, and false negatives; higher values indicate better event discrimination.
\begin{equation}
\mathrm{HSS} = \frac{2\,(TP \times TN - FP \times FN)}{(TP + FN)\,(FN + TN) + (TP + FP)\,(FP + TN)}
\end{equation}
In this paper, we use a precipitation threshold \(th = 0.1\,\mathrm{mm}\) to classify rain versus no-rain events.

\paragraph{Continuous Ranked Probability Score (CRPS)} is estimated for our ensembles as follows:
\begin{equation}
    \mathrm{CRPS}_{t,h,w} = \frac{1}{M} \sum_{m=1}^{M} |\hat{y}_{m,t,h,w} - y_{t,h,w}| - \frac{1}{2M^2} \sum_{m=1}^{M} \sum_{n=1}^{M} |\hat{y}_{m,t,h,w} - \hat{y}_{n,t,h,w}|
    \label{equ:crps_ensemble}
\end{equation}
where $M=50$ for all compared systems and $\hat{y}_{m,t,h,w}$ is the $m$-th ensemble member at a given case and grid point. Following~\citet{price2025probabilistic}, we use the traditional estimator under the matched ensemble size. Gridpoint CRPS values are averaged over the evaluation cases and spatial domain using the latitude weights in Eq.~\ref{equ:Li}.

\begin{figure}[t]
    \centering
    
    \includegraphics[width=0.62\linewidth]{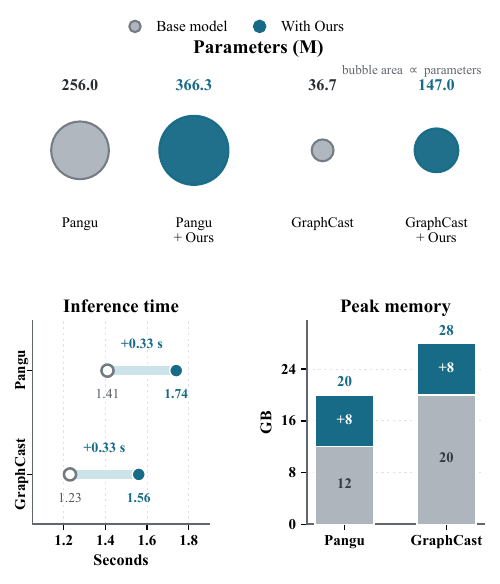} 
    
    \caption{Computational overhead of our framework on Pangu-Weather and GraphCast in terms of parameters, inference time, and peak GPU memory. Pangu-Weather parameter counts include all four forecast models.}
    \label{fig:complexity}

\end{figure}

\paragraph{Brier Score (BS)}
\label{appendix:bs}
BS measures the mean squared difference between predicted probabilities and binary outcomes for precipitation events. Given an ensemble of $M$ members,
\[
\hat{Y} = \{\hat{y}^{(1)}, \hat{y}^{(2)}, \dots, \hat{y}^{(M)}\},
\]
and a specific rainfall threshold $th$, we first define the empirical predictive probability $P$ at each grid point and time step as the fraction of members exceeding the threshold:
\begin{equation}
P = \frac{1}{M} \sum_{m=1}^{M} \mathbb{I}(\hat{y}^{(m)} \geq th)
\end{equation}
where $\mathbb{I}(\cdot)$ is the indicator function. The Brier Score is then calculated by averaging the squared error between this probability and the binary observation $O = \mathbb{I}(y \geq th)$:
\begin{equation}
\mathrm{BS} = \frac{1}{T \cdot H \cdot W} \sum_{t,h,w} L(h)(P_{t,h,w} - O_{t,h,w})^2
\label{equ:bs_code_logic}
\end{equation}
where $T$, $H$, and $W$ denote the number of evaluated cases, grid height, and grid width. BS ranges from 0 to 1, with 0 indicating perfect event-probability accuracy; we use it as an aggregate skill score rather than attributing changes separately to calibration or resolution. For BS, the precipitation threshold is set to $0.1\mathrm{mm}$.

\begin{figure}[t]
    \centering
    \includegraphics[width=0.65\linewidth]{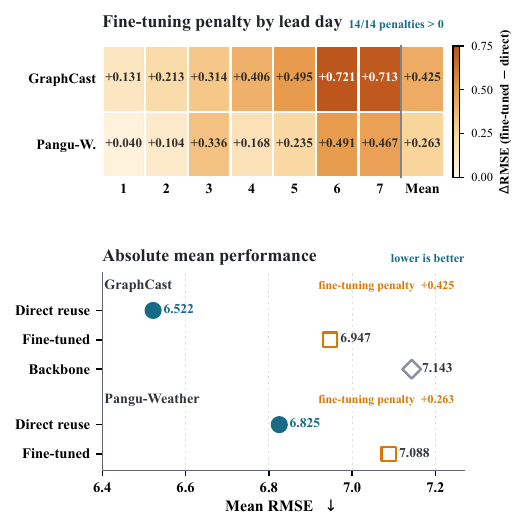} 
    
    \caption{Cross-backbone reuse and fine-tuning diagnostic.}
    \label{fig:cross_backbone_finetuning_diagnostic}
    
\end{figure}

\subsection{Additional Results}

\paragraph{Complexity and Efficiency Analysis}
Figure~\ref{fig:complexity} reports the computational overhead of the proposed generator. It introduces $110.3$M parameters, $0.33$ s per generated sample, and approximately $8$ GB of additional peak GPU memory under the reported configuration, making the relative overhead more pronounced for lightweight backbones. This non-negligible cost mainly arises from iterative diffusion sampling. However, the generator is trained once and reused across different forecast backbones at the same resolution, avoiding backbone-specific retraining, and fewer sampling steps can be used when lower latency is preferred. Together with the sampling-step analysis, these results expose the trade-off between precipitation quality and computational cost.

\paragraph{Temporal-Density Ablation.}
\label{appendix:temporal}
We compare the default two-state input $(\bm{\hat{X}}_{t-12},\bm{\hat{X}}_t)$ with four states sampled at 6-hour intervals. The two configurations achieve mean RMSEs of 6.520 and 6.518, respectively, compared with 7.143 for native GraphCast. Denser conditioning reduces RMSE by only 0.002 (0.03\%), while the two-state input reduces inference overhead from 0.84\,s to 0.33\,s under the evaluated configuration. We therefore adopt two-state conditioning for its lower computational cost and similar forecast accuracy.

\begin{figure}[t]
    \centering
    \includegraphics[width=0.62\linewidth]{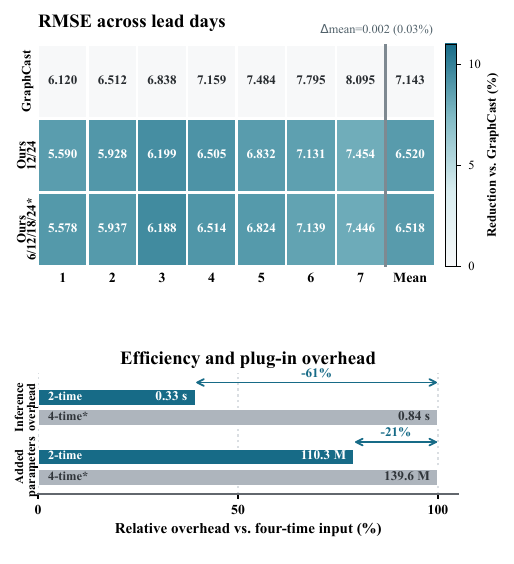} 
    
    \caption{Temporal-density ablation under GraphCast conditioning: forecast accuracy and computational overhead with two-state and four-state atmospheric inputs.}
    \label{fig:temporal_conditioning_ablation}
    
\end{figure}

\paragraph{Cross-Backbone Reuse and Fine-Tuning Diagnostic}
\label{app:cross_backbone}
At a fixed resolution, GraphCast and Pangu-Weather are coupled with the same ERA5-trained generator checkpoint and use an identical interface in terms of variables, grids, units, and normalization. This controlled setting isolates the effect of the generator adaptation strategy from differences in model configuration. Direct reuse applies the pretrained generator to forecast states without parameter updates, whereas the diagnostic fine-tuning variant starts from the same checkpoint and updates only the generator using forecast-state--IMERG pairs; the forecast backbone remains frozen.

As shown in Figure~\ref{fig:cross_backbone_finetuning_diagnostic}, fine-tuning increases RMSE at every evaluated lead day for both backbones. The mean RMSE increases from 6.522 to 6.947 for GraphCast, corresponding to a fine-tuning penalty of 0.425, and from 6.825 to 7.088 for Pangu-Weather, corresponding to a penalty of 0.263. The degradation becomes particularly pronounced at longer lead times, where forecast errors and their lead-time dependence are stronger. Moreover, direct reuse with GraphCast achieves a substantially lower mean RMSE than the original GraphCast forecast (6.522 versus 7.143). These consistent results suggest that the ERA5-trained checkpoint captures a transferable atmosphere--precipitation relationship, while fitting the generator to backbone-specific forecast errors may cause it to absorb lead-time-dependent biases and consequently weaken this learned relationship. We therefore adopt direct cross-backbone reuse in the primary method and include fine-tuning only as a diagnostic experiment.

\begin{figure}[!htbp]
    \centering
    \includegraphics[width=1\textwidth]{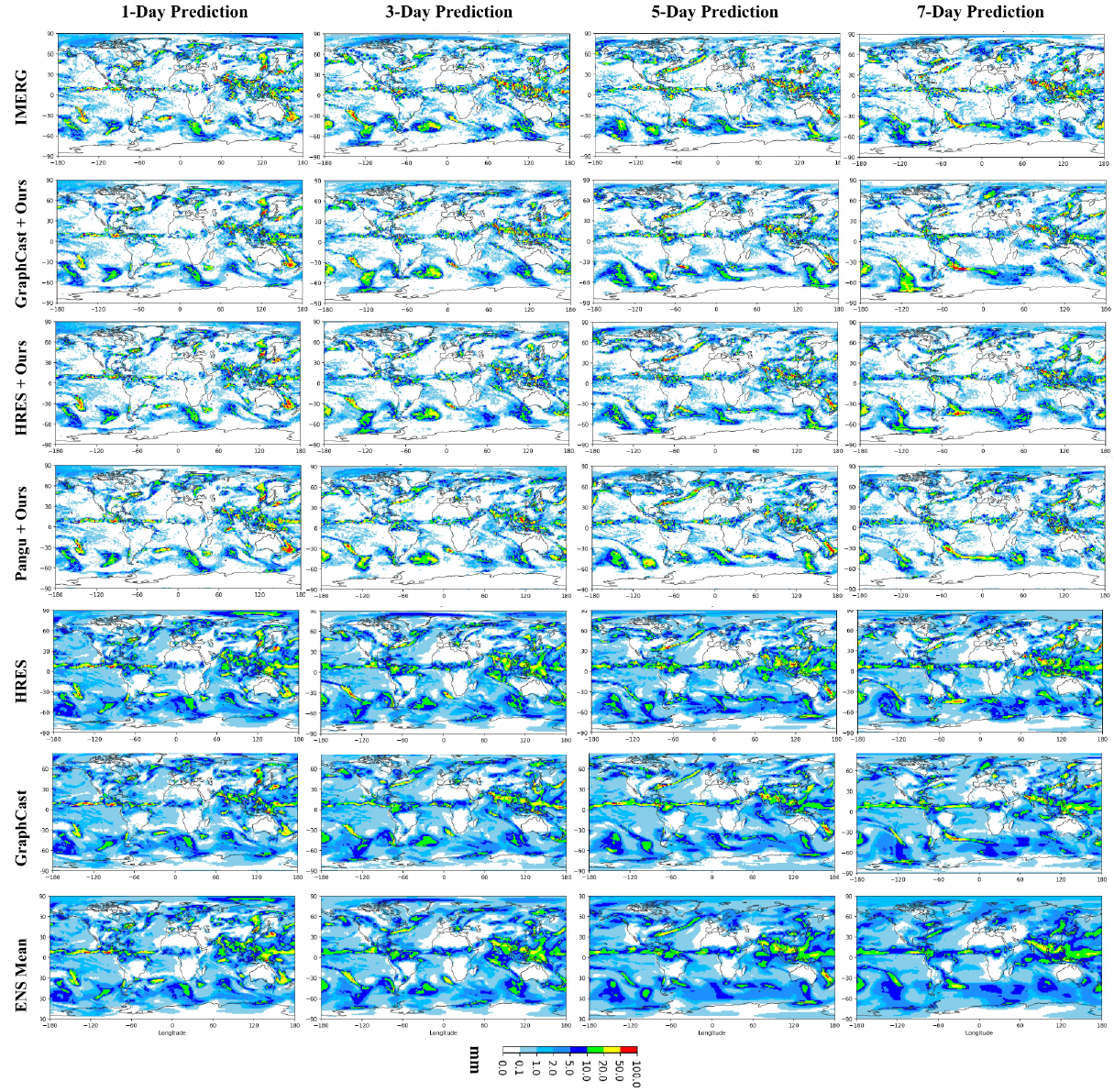} 
    \caption{Global precipitation fields at lead times of 1, 3, 5, and 7 days, with IMERG as the reference. Forecasts are initialized at 00:00 UTC on July 6, 2022.}
    \label{fig:all_visual}
    
\end{figure}

\paragraph{Additional global visualization comparison.} We provide additional visualization comparisons for all models at lead times of 1, 3, 5, and 7 days in Figure~\ref{fig:all_visual}.

\begin{figure}[t]
    \centering
    \includegraphics[width=1\textwidth]{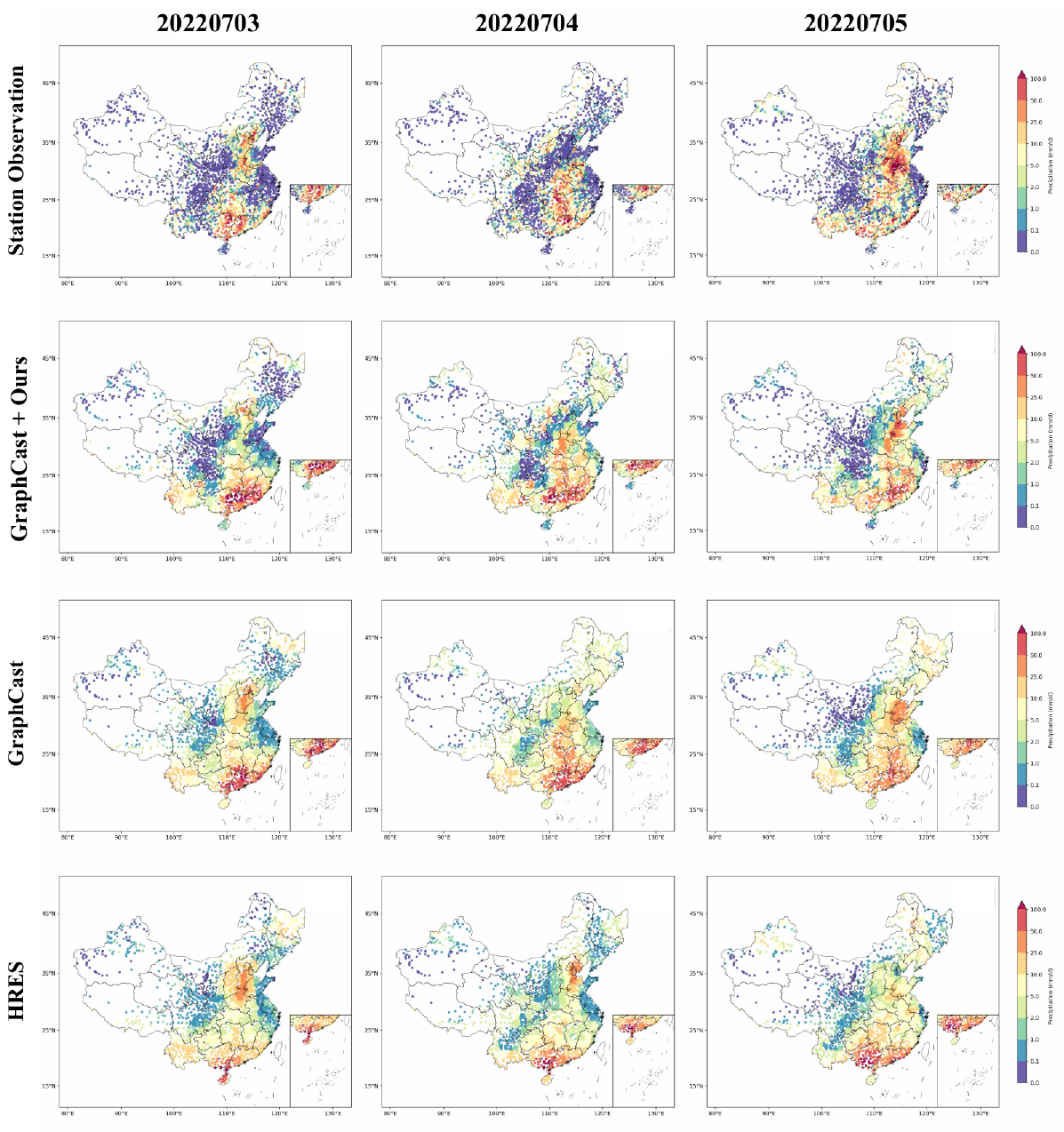}
    \caption{Station-based comparison across three valid times, with forecasts initialized on June 29, 2022.}
    \label{fig:all_sta_compare}
    
\end{figure}

\paragraph{Additional Station Observation Comparison.} We have added more site observation visualizations and comparisons at different time points in Figure~\ref{fig:all_sta_compare}.

\end{document}